\documentclass[conference]{IEEEtran}
\IEEEoverridecommandlockouts
\usepackage{cite}
\usepackage{amsmath,amssymb,amsfonts}
\usepackage{graphicx}
\usepackage{textcomp}
\usepackage{xcolor}

\usepackage{algorithm}
\usepackage{array}
\usepackage{microtype}
\usepackage{graphicx}
\usepackage{subfigure}
\usepackage{booktabs} 
\usepackage{amssymb}
\usepackage{mathtools}
\usepackage{amsthm}
\usepackage{amsfonts}       
\usepackage{nicefrac}       
\usepackage{microtype}      
\usepackage{fancyhdr}       
\usepackage{graphicx}       
\usepackage{listings}
\usepackage{multirow}
\usepackage{lineno}
\usepackage{bm}
\usepackage{amsmath,amssymb,amsfonts}
\usepackage{amsmath}
\usepackage{algpseudocode}
\usepackage{listings}
\usepackage{url}

\def\BibTeX{{\rm B\kern-.05em{\sc i\kern-.025em b}\kern-.08em
    T\kern-.1667em\lower.7ex\hbox{E}\kern-.125emX}}
\begin{document}

\title{Data-free Knowledge Distillation \\ with Diffusion Models\thanks{*Corresponding author: Jun Yu, junyu@ustc.edu.cn}}


\author{
\IEEEauthorblockN{
Xiaohua Qi\textsuperscript{1},
Renda Li\textsuperscript{1},
Long Peng\textsuperscript{1},
Qiang Ling\textsuperscript{1},
Jun Yu\textsuperscript{1}\textsuperscript{*},
Ziyi Chen\textsuperscript{2},
Peng Chang\textsuperscript{2},
Mei Han\textsuperscript{2}
Jing Xiao\textsuperscript{2}
}

\IEEEauthorblockA{\textsuperscript{1}University of Science and Technology of China, Hefei, China}

\IEEEauthorblockA{\textsuperscript{2}PAII Inc.}
}

\maketitle

\begin{abstract}
Recently Data-Free Knowledge Distillation (DFKD) has garnered attention and can transfer knowledge from a teacher neural network to a student neural network without requiring any access to training data. Although diffusion models are adept at synthesizing high-fidelity photorealistic images across various domains, existing methods cannot be easiliy implemented to DFKD. To bridge that gap, this paper proposes a novel approach based on diffusion models, DiffDFKD. Specifically, DiffDFKD involves targeted optimizations in two key areas. Firstly, DiffDFKD utilizes valuable information from teacher models to guide the pre-trained diffusion models' data synthesis, generating datasets that mirror the training data distribution and effectively bridge domain gaps. Secondly, to reduce computational burdens, DiffDFKD introduces Latent CutMix Augmentation, an efficient technique, to enhance the diversity of diffusion model-generated images for DFKD while preserving key attributes for effective knowledge transfer. Extensive experiments validate the efficacy of DiffDFKD, yielding state-of-the-art results exceeding existing DFKD approaches.We release our code at \url{https://github.com/xhqi0109/DiffDFKD}.
\end{abstract}


\section{Introduction}
Knowledge distillation (KD) ~\cite{fang2022up} has emerged as a prominent technique for model compression, enabling the transfer of knowledge from a large and highly accurate teacher model to a smaller and more efficient student model. The key insight of KD is to leverage the rich representation knowledge encoded within the teacher's outputs and activations as a supervisor for training the student. By transcending simple one-hot labels and mimicking the teacher's softer logit distributions, the student can acquire a more nuanced understanding of the task, ultimately achieving better performance than training from scratch. KD has found widespread applications across domains, facilitating the deployment of compact models on resource-constrained devices while retaining remarkable accuracy~\cite{yin2019dreaming}.

Although traditional KD approaches have achieved considerable success~\cite{chen2019data, fang2022up}, a major limitation arises when the original training data is unavailable or inaccessible due to privacy, security, or proprietary concerns~\cite{fang2022up}. In such scenarios, data-free knowledge distillation emerges as a solution, enabling knowledge transfer without direct access to the teacher's training data~\cite{fang2022up}. Typically, DFKD methods use a ``distill-by-generate'' approach, focusing on model inversion to recreate a dataset based on the internal representations and decisions of the pre-trained teacher model~\cite{yin2020dreaming}. Once this synthetic data are generated, conventional data-driven KD techniques can be leveraged to train the student model, effectively circumventing the need of the original training data.

A primary challenge within DFKD lies in both the quality and diversity of the synthetic data~\cite{fang2022up, choi2020data}. Prior methods~\cite{fang2021contrastive, yin2020dreaming, chen2019data, yu2023data, liu2024small} often commence with the generator initialized by random noise, neglecting the incorporation of meaningful semantic priors or valuable information from other proficient pre-trained models with robust generative capabilities. This neglect frequently culminates in synthetic samples that lack natural image statistics~\cite{yin2020dreaming}, thereby significantly impeding the knowledge transfer process. Furthermore, a recurring issue is mode collapse, where the generated samples exhibit a lack of diversity, converging to similar modes within the data distribution~\cite{yin2020dreaming, chen2019data}. Consequently, the student model, trained on this limited and homogeneous synthetic data, fails to develop a comprehensive understanding of the task, leading to suboptimal performance outcomes.

Recently, diffusion models~\cite{rombach2022high} have emerged as a powerful class of generative models, demonstrating an unprecedented capability to synthesize high-fidelity photorealistic images across a wide variety of domains. Unlike traditional generative adversarial networks (GANs) ~\cite{fang2022up}, which often suffer from training instability and mode collapse, diffusion models learn to gradually denoise a signal from pure noise, enabling more diverse and semantically coherent image generation. While the impressive generative capabilities of diffusion models have been primarily explored for tasks such as super-resolution~\cite{peng2025towards} and few-shot classification~\cite{zhang2023federated}, their potential impact on DFKD remains largely unexplored.

Leveraging the generative power of diffusion models could pave the way for more effective and scalable DFKD approaches, overcoming the limitations of existing synthesis-based methods~\cite{fang2021contrastive, fang2022up, yin2020dreaming, chen2019data}. However, current methods fall short in fully realizing the potential of diffusion models. For instance, DM-KD~\cite{li2023synthetic} integrates a pre-trained diffusion model with label annotations from a teacher model to synthesize data for DFKD, achieving state-of-the-art performance on datasets like CIFAR-100~\cite{krizhevsky2009learning}. Nonetheless, DM-KD overlooks the potential of leveraging richer information within the teacher model, such as statistical moments from batch normalization layers. Consequently, this leads to a notable distributional discrepancy between synthesized data and original training data, rendering DM-KD ineffective across various specialized domains like QuickDraw~\cite{peng2019moment}. Moreover, in scenarios where access is limited solely to the teacher model without accompanying label annotations~\cite{fang2022up}, DM-KD fails to facilitate domain adaptation, restricting its effectiveness to the original training domains of the diffusion model.

Motivated by these limitations and the immense potential of diffusion models, we propose DiffDFKD, a novel DFKD framework that harnesses pre-trained diffusion models for data synthesis while effectively leveraging knowledge from teacher models. Our DiffDFKD bridges domain gaps by guiding the diffusion model's synthesis process with information from teacher models. Specifically, we leverage valuable information from teacher models to optimize the latent representations, generating domain-customized data to closely mirror the original distribution. Additionally, to reduce computational burdens, we introduce Latent CutMix Augmentation (LCA), a computationally efficient technique that enhances the diversity of synthesized images while preserving essential characteristics for effective knowledge transfer. By exploiting diffusion models' powerful synthesis capabilities and utilizing teacher model knowledge, DiffDFKD ensures optimal knowledge distillation tailored to diverse domains, outperforming conventional DFKD methods. The main contributions of this paper can be summarized as follows.

\begin{itemize}
    \item We propose a novel data-free knowledge distillation framework, named DiffDFKD, which comprises two stages. At the first stage, we guide the diffusion model's synthesis process using the teacher's pre-trained model to generate domain-customized data. At the second stage, we utilize the synthesized data to perform knowledge distillation.
    \item We propose Latent CutMix Augmentation, a novel and computationally efficient technique that increases the diversity of diffusion model generated images for data-free knowledge distillation while preserving characteristics essential for successful knowledge transfer.
    \item Extensive experiments validate the efficacy of DiffDFKD, yielding state-of-the-art results exceeding existing DFKD approaches.
\end{itemize}

\section{Related Work}
\subsection{Data-Free Knowledge Distillation}

Data-Free Knowledge Distillation~\cite{yin2019dreaming, fang2021contrastive} aims to train a compact student model from a pre-trained teacher model without direct access to the original training data. Traditional approaches predominantly use GANs or noise optimization-based synthetic data generation within the "distilling-by-generating" paradigm. For example, ZSKT~\cite{micaelli2019zero} employs adversarial training to bridge disparities between student and teacher models using the KL divergence~\cite{fang2022up}. Chen et al.~\cite{chen2019data} proposed DFAD, which uses Mean Absolute Error to address gradient decay issues, while Choi et al.~\cite{choi2020data} introduced DFQ for model quantization, utilizing batch categorical entropy maximization to ensure class representation. Despite these advancements, challenges remain in terms of prolonged training times and large batch sizes. To overcome this, Fast~\cite{fang2022up} employs a meta-generator for accelerated knowledge transfer. Additionally, SSD-KD~\cite{liu2024small} incorporates reinforcement learning for data synthesis, and the NAYER~\cite{tran2024nayer} method relocates noise to a layer while using label-text embeddings to enhance sample quality and training speed. Despite these innovations, many existing methods still generate low-quality, limited-diversity images. The generative capabilities of diffusion models offer a promising avenue for DFKD, yet their potential remains largely unexplored.

\begin{figure*}[t]
    \centering

    \includegraphics[width=\linewidth]{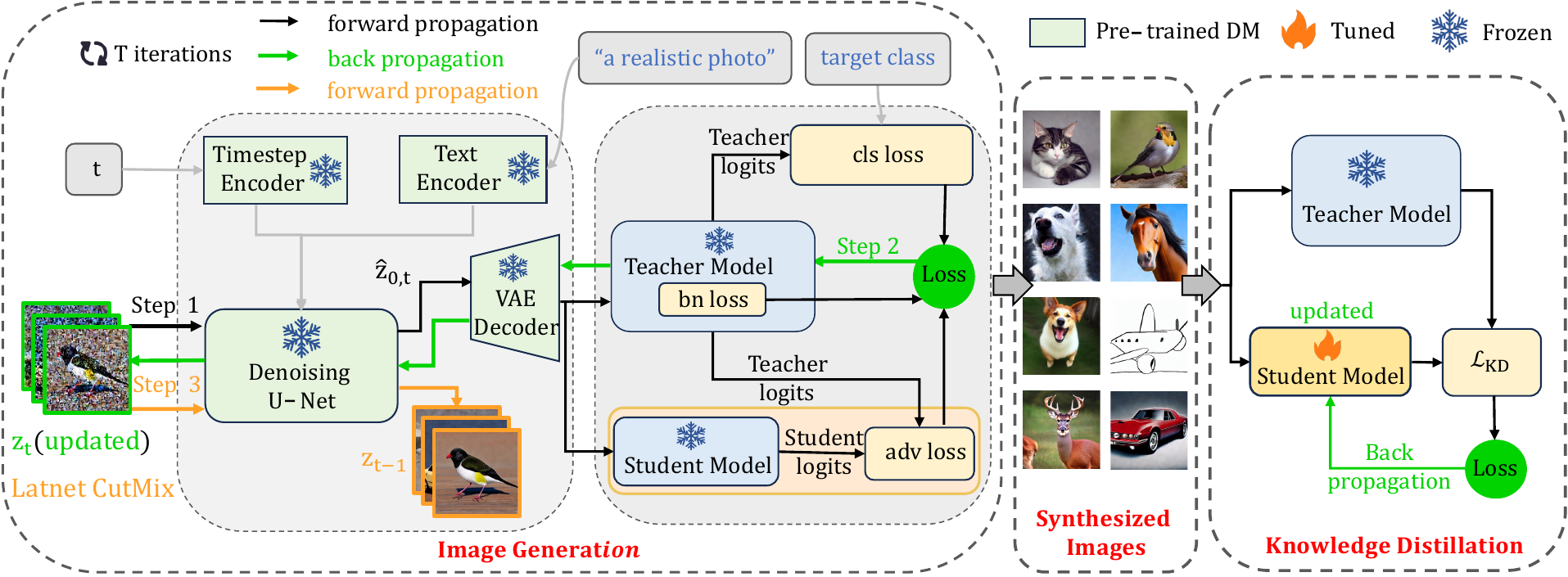}
    \vspace{-5mm}

    \caption{Overview of the proposed DiffDFKD framework. At each step $t$: (1) the latent variable $\mathbf{z}_t$ is processed through a pre-trained diffusion model to obtain a prediction $\hat{\mathbf{z}}_{0,t}$, which is used to compute the loss; (2) the combined loss is backpropagated to update $\mathbf{z}_t$; (3) the updated $\mathbf{z}_t$ undergoes CutMix augmentation and denoising to produce $\mathbf{z}_{t-1}$ for the next iteration. After $T$ steps, a synthetic dataset is generated, facilitating the knowledge distillation process. Note that in the figure, the Image Generation and Knowledge Distillation stages utilize the same teacher-student model pair.}
    \vspace{-5mm}

    \label{fig:framework}
\end{figure*}

\section{Methodology}

The overall procedure of our proposed DiffDFKD is depicted in Figure~\ref{fig:framework}. We first provide essential preliminaries on diffusion models in Section~\ref{sec:preliminaries}. Then  Section~\ref{sec:Inversion_Loss } presents how to constrain Diffusion's data synthesis to minimize the difference between synthetic and real data distributions. Subsequently, Section~\ref{sec:edited_latent} elaborates on the application of constraints to the Diffusion data synthesis process. To enhance computational efficiency and diversify the generated data, we introduce Latent Cutmix Augmentation in Section~\ref{sec:lca}. Finally, Section~\ref{sec:data-free-kd} describes how we utilize the generated data for DFKD.

\subsection{Preliminaries}
\label{sec:preliminaries}

The diffusion model~\cite{rombach2022high} $\epsilon_{\theta}$ samples initial noise $\mathbf{x}_T$ from a standard Gaussian distribution $\mathcal{N}(\mathbf{0}, \mathbf{I})$ and then generates realistic images $\mathbf{x}_0$ through an iterative denoising process, where $T$ denotes the maximum number of iterations. To reduce the computational cost, Latent Diffusion Models (LDMs)~\cite{rombach2022high} transform the image space $\mathbf{x}$ into the latent space $\mathbf{z}$ through an encoder and decoder Variational Auto-Encoder (VAE)~\cite{rombach2022high}. In the latent space, the reverse diffusion process is expressed as:
\begin{equation}
p{(\mathbf{z}_{t-1}|\mathbf{z}_{t})} = \mathcal{N}(\mu_\theta(\mathbf{z}_t,t), \Sigma_\theta(\mathbf{z}_t,t)),
\label{eq:ldf_3}
\end{equation}
where $\mathbf{z_t}$ is the noisy latent representation at timestep $t$, and $\mu_\theta$ and $\Sigma_\theta$ represent the predicted mean and variance of the model, respectively.

To control the content of the generated images, LDMs, and other models introduce a conditional input $c$, allowing the noise prediction to be conditioned on $c$ as:
\begin{equation}
p{(\mathbf{z}_{t-1}|z_t, c)} = \mathcal{N}(\mu_\theta(\mathbf{z}_{t},t|c), \Sigma_\theta(\mathbf{z}_{t},t|c)).
\label{eq:ldf_4}
\end{equation}

To generate images with distinctive features being more related to the given condition $c$, a classifier-free sampling strategy~\cite{rombach2022high} is employed. By using a noise prediction network $\epsilon_\theta$ to predict noise, the denoising process can be guided toward removing the conditional noise as:
\begin{equation}
\hat{\epsilon}_\theta(\mathbf{z}_t, t|c) = \epsilon_\theta(\mathbf{z}_t, t|\emptyset) + s \cdot (\epsilon_\theta(\mathbf{z}_t, t|c)-\epsilon_\theta(\mathbf{z}_t, t|\emptyset)),
\label{eq:epsilon}
\end{equation}
where $\epsilon_\theta(\mathbf{z}_t,t|c)$ is the predicted noise at timestep $t$ given condition $c$, $\epsilon_\theta(\mathbf{z}_t,t|\emptyset)$ is the unconditional predicted noise, the hyperparameter $s\geq1$ adjusts the guidance strength, $s=1$ indicates no classifier-free guidance is employed, and $\emptyset$ represents a trainable ``null'' condition.

At any timestep $t$, the diffusion model predicts noise $\hat{\epsilon}_\theta(\mathbf{z}_t, t|c)$, which is then used to predict the denoised latent variable $\hat{\mathbf{z}}_{0,t}$. This prediction is given by the following:
\begin{linenomath}
\begin{align}
    \hat{\mathbf{z}}_{0,t} = \frac{\mathbf{z}_{t} - \sqrt{1 - \alpha_{t}} \hat{\epsilon}_{\theta}(\mathbf{z}_{t}, t|c)}{\sqrt{\alpha_{t}}}.
    \label{eq:get_z0}
\end{align}
\end{linenomath}

Subsequently, leveraging $\hat{\mathbf{z}}_{0,t}$ obtained from~\eqref{eq:get_z0}, the latent variable for the next timestep, $\mathbf{z}_{t-1}$, can be determined as:
\begin{linenomath}
\begin{align}
    \mathbf{z}_{t-1} = \sqrt{\alpha_{t-1}} \cdot \hat{\mathbf{z}}_{0,t} + \sqrt{1 - \alpha_{t-1}} \cdot \varepsilon_t,
    \label{eq:get_z_t-1}
\end{align}
\end{linenomath}
where, $\alpha_{t}$, $\alpha_{t-1}$ are pre-defined parameters of the diffusion schedule, and $\varepsilon_t$ represents Gaussian noise sampled at each timestep.

\subsection{Loss Function for Data Generation}
\label{sec:Inversion_Loss }

In this section, we describe strategies to employ the pre-trained teacher model $f_t(x; \theta_t)$ to guide the diffusion model in synthesizing images that match the teacher's domain distribution. Our aim is to reconstruct training data $\mathcal{D^\prime}$ from $f_t(x; \theta_t)$, serving as an alternative to the inaccessible original data $\mathcal{D}$ for DFKD.

\paragraph{Batch Normalization Regularization} Originally introduced by~\cite{yin2020dreaming}, this technique efficiently utilizes the teacher model's statistical moments from batch normalization layers. The regularization is typically represented as the divergence between the feature statistics $\mathcal{N}(\mu_l(x), \sigma_l^2(x))$ and batch normalization statistics $\mathcal{N}(\mu_{l}, \sigma^2_{l})$, 
\begin{equation}
   \mathcal{L}_{bn}(x) = \sum_{l} D\left( \mathcal{N}(\mu_l(x), \sigma_l^2(x)), \mathcal{N}(\mu_{l}, \sigma^2_{l}) \right).
\end{equation}
\paragraph{Class Prior} Typically introduced for class-conditional generation, this approach puts a ``one-hot'' assumption on the network predictions of $x\in \mathcal{D^\prime}$~\cite{chen2019data}. Given a pre-defined category $y$, it encourages to minimize the following cross-entropy loss: 
\begin{equation}
   \mathcal{L}_{cls}(x) = CE(f_t(x), y).
\end{equation}
\paragraph{Adversarial Distillation} Motivated by robust optimization, this technique forces $x$ to produce large disagreement between the teacher $f_t(x; \theta_t)$ and student $f_s(x; \theta_s)$~\cite{micaelli2019zero, fang2022up}, i.e., maximizing the following KL divergence loss:
\begin{equation}
   \mathcal{L}_{adv}(\mathbf{x}) = - KL(f_t(\mathbf{x}) / \tau \| f_s(\mathbf{x}) / \tau).
   \label{eqn:adv}
\end{equation}
\paragraph{Unified Framework} Combining the aforementioned techniques leads to a unified inversion loss~\cite{choi2020data} for DFKD as:
\begin{equation}
   \mathcal{L}_{inv}(x) = \alpha \cdot \mathcal{L}_{bn}(x) + \beta \cdot \mathcal{L}_{cls}(x) + \gamma \cdot \mathcal{L}_{adv}(x), \label{eqn:unified}
\end{equation}
where $\alpha$, $\beta$, and $\gamma$ are balancing weights for different criteria.

\subsection{Diffusion Denoising with Edited Latent}
\label{sec:edited_latent}
Traditional classifier-free sampling methods, based on conditional text prompts, null conditioning, and scale factors, often lead to distributional inconsistencies between $\mathcal{D^\prime}$ and $\mathcal{D}$. One approach to mitigating this issue involves fine-tuning the DMs parameters using the loss function in~\eqref{eqn:unified}. However, this method is computationally expensive and can significantly degrade the original generative capabilities of the DMs. Alternatively, inspired by~\cite{yang2023one}, another approach involves editing the initial noise $\mathbf{z}_{T}$. However, updating $\mathbf{z}_{T}$ is both memory and computationally expensive, as it requires iterative inference over $T$ steps.

To address these challenges, we propose a method for updating the latent $\mathbf{z}_{t}$. This method only requires a single inference per update, which is computationally efficient. Specifically, at any timestep $t$, we first use~\eqref{eq:get_z0} to obtain the predicted $\hat{\mathbf{z}}_{0,t}$. Then, we compute the loss function as defined in~\eqref{eqn:unified} and its gradient to update $\mathbf{z}_{t}$, adjusting the latent variable over $t$ steps as:
\begin{linenomath}
\begin{align} 
\hat{\mathbf{z}}_t = \mathbf{z}_t - \eta \nabla_{\mathbf{z}_t} \mathcal{L}_{inv}(\hat{\mathbf{z}}_{0,t}).
\label{eq:update_zt}
\end{align}
\end{linenomath}

Although $\hat{\mathbf{z}}_{0,t}$ may initially appear blurry in the early timesteps, the noise level is significantly reduced compared to $\mathbf{z}_{t}$. Unlike the approach in~\cite{yang2023one}, our method optimizes the latent variable $\hat{\mathbf{z}}_{0,t}$ instead of iteratively updating the initial noise $\mathbf{z}_{T}$ to obtain $\mathbf{z}_0$. This reduction in memory and computational requirements significantly enhances computational efficiency while satisfying the needs for DFKD.

After updating $\hat{\mathbf{z}}_{t}$, we use~\eqref{eq:get_z_t-1} to iteratively denoise and obtain $\mathbf{z}_{t-1}$, eventually leading to a partially synthesized but realistic $\mathbf{z}_0$. Subsequently, we utilize the VAE decoder to convert $\mathbf{z}_0$ into a realistic image $\mathbf{x}_0$.

\algdef{SE}[SUBALG]{Indent}{EndIndent}{}{\algorithmicend\ }%
\algtext*{Indent}
\algtext*{EndIndent}

\algnewcommand{\LineComment}[1]{\State {\(//\) #1}}

\begin{table*}[]
\centering
\vspace{-4mm}
\caption{DFKD results on CIFAR-10, CIFAR-100, Tiny-ImageNet, and DomainNet. The best-performing method is highlighted in bold. In this 
table, `R' represents ResNet, `W' corresponds to WideResNet, and `V' stands for VGG.}
\vspace{-2mm}
\begin{tabular}{cccccccccc}
\toprule
\multicolumn{1}{l}{\textbf{}}    & \multicolumn{5}{c}{\textbf{CIFAR10}}                                               & \multicolumn{2}{c}{\textbf{CIFAR100}} & \textbf{TinyImageNet} & \textbf{DomainNet} \\
\hline
\multirow{2}{*}{\textbf{Method}} & \textbf{R34}   & \textbf{V11}   & \textbf{W40-2} & \textbf{W40-2} & \textbf{W40-2} & \textbf{R34}      & \textbf{V11}      & \textbf{R34}          & \textbf{R34}       \\
                                 & \textbf{R18}   & \textbf{R18}   & \textbf{W16-1} & \textbf{W16-2} & \textbf{W40-1} & \textbf{R18}      & \textbf{R18}      & \textbf{R18}          & \textbf{R18}       \\
                                 \hline
Teacher                          & 95.70          & 92.25          & 94.87          & 94.87          & 94.87          & 78.05             & 71.32             & 66.44                 & 62.26              \\
Student                          & 95.20          & 95.20          & 91.12          & 93.94          & 93.95          & 77.10             & 77.10             & 64.87                 & 63.69              \\
DeepInv                          & 93.26          & 90.36          & 83.04          & 86.85          & 89.72          & 61.32             & 54.13             & -                     & -                  \\
CMI                              & 94.84          & 91.13          & 90.01          & 92.78          & 92.52          & 77.04             & 70.56             & 64.01                 & -                  \\
DAFL                             & 92.22          & 81.10          & 65.71          & 81.33          & 81.55          & 74.47             & 54.16             & -                     & -                  \\
ZSKT                             & 93.32          & 89.46          & 83.71          & 86.07          & 89.66          & 67.74             & 54.31             & -                     & -                  \\
DFQ                              & 94.61          & 90.84          & 86.14          & 91.69          & 92.01          & 77.01             & 66.21             & 63.73                 & -                  \\
Fast                             & 94.05          & 90.53          & 89.29          & 92.51          & 92.45          & 74.34             & 67.44             & -                     & 54.56              \\
SpaceshipNet                     & 95.39          & 92.27          & 90.38          & 93.25          & 93.56          & 77.41             & 71.41             & 64.04                 & -                  \\
SSD-KD                           & 94.26          & 90.67          & 89.96          & 93.11          & 93.23          & 75.16             & 68.77             & -                     & -                  \\
DM-KD                            & 93.78          & 89.69          & 87.65          & 92.21          & 92.16          & 76.14             & 70.43             & 63.61                 & 45.39              \\
Ours                             & \textbf{95.41} & \textbf{92.36} & \textbf{91.10} & \textbf{94.03} & \textbf{93.81} & \textbf{77.43}    & \textbf{71.76}    & \textbf{64.11}        & \textbf{61.19}     \\
\bottomrule
\end{tabular}
\vspace{-3mm}

\label{tbl:benchmark_cifar10}

\end{table*}

\subsection{Latent CutMix Augmentation}
\label{sec:lca}

Due to the computationally intensive nature of the diffusion model's generative process, which requires iterative denoising steps (e.g., 50 steps) to obtain a realistic image $\mathbf{x}_{0}$, we explore the potential of using intermediate latent representations $\hat{\mathbf{z}}_{t}$. Unfortunately, these intermediate representations often exhibit significant similarity. To address this, we propose a novel LCA technique to enhance the model's ability to generate diverse images. Specifically, we apply CutMix augmentation to the updated latent representations $\hat{\mathbf{z}}_t$ from~\eqref{eq:update_zt} during the iterative denoising process.

Following traditional CutMix~\cite{yun2019cutmix}, we use the following operation to augment the latent representations $\hat{\mathbf{z}}_t$:
\begin{equation}
    \vspace{-1mm}
    \hat{\mathbf{z}}_t = \psi(\hat{\mathbf{z}}_t),
    \vspace{-1mm}
    \label{eq:cutmix}
\end{equation}
where $\hat{\mathbf{z}}_t$ represents the updated latent representations obtained from~\eqref{eq:update_zt} for a batch of data and $\psi$ denotes the CutMix augmentation operation. To enhance the variety of generated images, we randomly select two latent codes from $\hat{\mathbf{z}}_t$ and apply the CutMix operation with random intensity.

Although this method increases data diversity, the augmented data may contain artifacts. Directly using this data in distillation introduces noise, reducing student performance. To mitigate this, we do not apply LCA at every timestep $t$ but instead perform it every $k$ step. During these $k$ steps, the diffusion model's inpainting and repair capabilities are utilized, where regions with missing or distorted features are corrected, resulting in natural images.

For each iteration $i$, we perform iterative denoising for $T$ steps to obtain $\mathbf{z}_0$, the specified text prompt $c$, class label $y$, and the teacher's model $f_t(x; \theta_t)$ are provided. For each $t \in \{T-k, T-2k, \ldots, 0\}$, with $k$ being a hyperparameter for DFKD, the generated latent variable $\hat{\mathbf{z}}_{0,t}$ from~\eqref{eq:get_z0} is decoded by the VAE decoder. This decoding produces a real image $\hat{\mathbf{x}}_{0,t}$, denoted as ${\mathbf{x}^{t}_{i}}$, resulting in $({\mathbf{x}^{t}_{i}}, y)$. These pairs are then added to $\mathcal{D^\prime}_{i}$. After multiple denoising iterations of generation, we obtain the synthetic dataset:
\[
\mathcal{D^\prime}_{i} = \{({\mathbf{x}^{T-k}_{i}}, y), ({\mathbf{x}^{T-2k}_{i}}, y), \ldots, ({\mathbf{x}^{0}_{i}}, y)\}.
\]

Through multiple iterations, we can obtain the synthesized data $\mathcal{D^\prime} = \{\mathcal{D^\prime}_{i}\}_{i=1}^{N}$.

\begin{figure*}[t]
    \centering
    \includegraphics[width=\linewidth]{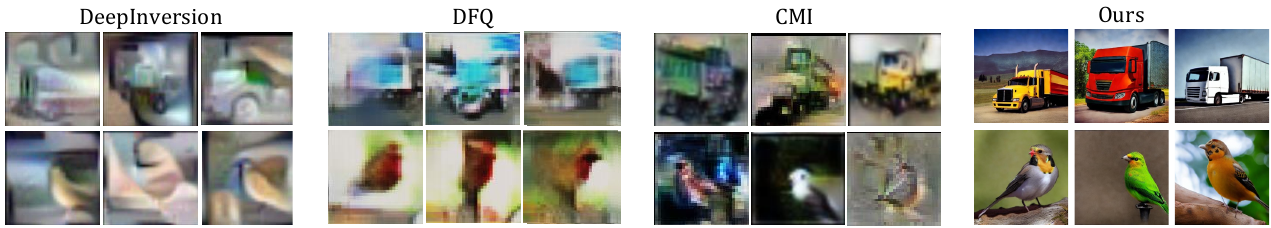}
    \vspace{-4mm}    
    \caption{Inverted data from a pre-trained ResNet-34 on CIFAR-10 with a student model of ResNet-18.}
    \label{fig:synthetic_samples}
    \vspace{-6mm}
\end{figure*}

\subsection{Data-Free Knowledge Distillation}
\label{sec:data-free-kd}

Although we leverage the denoising capabilities of DMs through $k$ steps to mitigate noise, some residual noise may still persist. To further reduce the distribution shift between $\mathcal{D^\prime}$ and $\mathcal{D}$, we adopt the $\mathcal{L}_{mSARC}$ loss proposed by~\cite{yu2023data} to address this issue. Specifically, $\mathcal{L}_{mSARC}$ constrains the class activation maps (CAMs)~\cite{zhou2016learning} of the feature at layer $i_k$ of the student network $f_s(x; \theta_s)$ to be consistent with CAMs of the feature at layer $j_k$ in the teacher network $f_t(x; \theta_t)$. This encourages the hidden layers at different stages of the student network to focus on the same spatial regions as the hidden layers of the teacher network at the corresponding stages. 

Additionally, we include the KL divergence loss as part of the knowledge distillation loss function:
\begin{equation}
    \mathcal{L}_{KD} = \lambda_{KL} \mathcal{L}_{mSARC} + \lambda_{mSARC} \mathcal{L}_{KL},
    \label{eq:update_student}
\end{equation}
where $\lambda_{KL}$ and $\lambda_{mSARC}$ are parameters that balance the two loss terms.

\section{Experiments}

\subsection{Settings}
\label{sec:exp_settings}
\paragraph{\textbf{Models and Datasets}}
In this paper, we evaluate the effectiveness of our proposed DiffDFKD framework across diverse network architectures, including ResNet~\cite{he2016deep}, VGG~\cite{simonyan2014very}, and Wide ResNet~\cite{zagoruyko2016wide}. We compare our approach with prevailing DFKD methods on four canonical classification datasets: CIFAR-10, CIFAR-100~\cite{krizhevsky2009learning}, Tiny-ImageNet~\cite{le2015tiny}, and DomainNet~\cite{peng2019moment}. CIFAR-10 and CIFAR-100 have a resolution of $32\times32$ pixels, Tiny-ImageNet has a resolution of $64\times64$ pixels, and DomainNet has a resolution of $224\times224$ pixels. DomainNet encompasses six diverse domains: Clipart, Infograph, Painting, QuickDraw, real, and Sketch. Due to the computational complexity associated with data generation, we follow~\cite{zhang2023federated} and limit our experiments on each domain to the initial ten categories. 

\paragraph{\textbf{Number of Synthesized Images}} To enhance generated image diversity, we employ DMs with text prompts "a realistic photo" plus random `subject' and `style' combinations. For the CIFAR-10 and CIFAR-100 datasets, we set $s=3$ and $step=10$, generating 15k images, with each image produced using four intermediate latent. For Tiny-ImageNet and DomainNet, we use the same settings, generating 30k and 12k images per domain, respectively. For the challenging QuickDraw domain, we employ $s=7.5$ and $step=70$.

\subsection{Benchmarks on CIFAR-10 and CIFAR-100}
Table~\ref{tbl:benchmark_cifar10} presents a benchmark comparison of existing DFKD on CIFAR-10 and CIFAR-100~\cite{krizhevsky2009learning} datasets. Our approach is juxtaposed with several baselines, including methods based on GANs or noise optimization: DeepInv~\cite{yin2020dreaming}, DAFL~\cite{chen2019data}, ZSKT~\cite{micaelli2019zero}, DFQ~\cite{choi2020data}, Fast~\cite{fang2022up}, SpaceshipNet~\cite{yu2023data}, and SSD-KD~\cite{liu2024small}. Additionally, we compare our method with DM-KD~\cite{li2023synthetic}, which is based on Diffusion. Due to the requirement of teacher label names for DM-KD~\cite{li2023synthetic}, we synthesized an equal amount of data using the same fixed prompts and DMs as our method for comparison purposes. For NAYER~\cite{tran2024nayer}, we did not conduct a direct comparison as it utilizes teacher label names. Our findings reveal that, compared to methods relying on GANs or noise optimization, our approach achieves state-of-the-art performance across all CIFAR-10 and CIFAR-100 model architectures. Furthermore, in contrast to the Diffusion-based DM-KD approach, our method demonstrates significant improvements across all datasets and model architectures.

Fig.~\ref{fig:synthetic_samples} visually demonstrates the synthetic data generated by DeepInversion~\cite{yin2020dreaming}, DFQ~\cite{choi2020data}, CMI~\cite{fang2021contrastive}, and our proposed method. The results illustrate that our approach excels in synthesizing data with richer semantic information, greater diversity, heightened realism, and enhanced local details, such as facial features and textures, compared to previous methods. This superiority translates to improved performance in the DFKD task. The visualized outcomes of DeepInversion~\cite{yin2020dreaming}, DFQ~\cite{choi2020data}, and CMI~\cite{fang2021contrastive} are sourced from~\cite{fang2021contrastive}. 

\begin{table}[]
\small 
\centering
    \caption{Student accuracy (\%) on $224\times224$ DomainNet. The teacher model is ResNet-34, and the student model is ResNet-18. T., S., Fast~\cite{fang2022up}, DM-KD~\cite{li2023synthetic}, and our method uses synthetic data. The results marked by ``*'' come from our re-implementations.}
\vspace{-2mm}

    \begin{tabular}{ccccccc}
  \toprule
        \textbf{Domain}    & \textbf{T.} & \textbf{S.} & \textbf{Fast}   & \textbf{DM-KD} & \textbf{Ours}  \\
        \midrule
        \textbf{Clipart}   & 67.21   & 70.65   & 60.72*    & 53.24* & \textbf{65.59} \\
        \textbf{Infograph} & 39.00   & 41.70   & 29.04*    & 35.68* & \textbf{39.21} \\
        \textbf{Painting}  & 43.99   & 48.88   & 31.83*    & 39.37* & \textbf{42.01} \\
        \textbf{QuickDraw} & 88.93   & 86.87   & 85.20*    & 23.80* & \textbf{88.78} \\
        \textbf{Real}      & 76.69   & 78.20   & 70.38*    & 72.29* & \textbf{76.95} \\
        \textbf{Sketch}    & 57.79   & 55.86   & 50.20*    & 48.00* & \textbf{54.62} \\
        \midrule
        \textbf{Avg}       & 62.26   & 63.69   & 54.56   & 45.39 & \textbf{61.19} \\
        \bottomrule
    \end{tabular}
    \vspace{-6mm}
\label{tbl:benchmark_domainnet}
\end{table}

\subsection{Benchmarks on Tiny-ImageNet and DomainNet}
To further validate the effectiveness of our method, we conduct experiments on higher-resolution datasets and additional domains. Table~\ref{tbl:benchmark_cifar10} presents the performance metrics on Tiny-ImageNet and the average domain performance on DomainNet. For Tiny-ImageNet, our method consistently outperforms other publicly available algorithms. The ``-'' symbol indicates that the corresponding algorithm has not been tested on this dataset. In the context of DomainNet, Table~\ref{tbl:benchmark_domainnet} provides a detailed breakdown of performance across different domains, comparing our method with Fast~\cite{fang2022up} and DM-KD~\cite{li2023synthetic}. Our method significantly outperforms Fast~\cite{fang2022up}, potentially due to our proposed DiffDFKD's ability to synthesize more realistic, diverse, and teacher model distribution-conforming data on larger resolution and challenging domains. In comparison to DM-KD~\cite{li2023synthetic}, we find it performs better than Fast~\cite{fang2022up} on the Infograph, Painting, and Real domains while exhibiting poorer performance on the Clipart, QuickDraw, and Sketch domains. DM-KD~\cite{li2023synthetic}, without utilizing teacher information, can only synthesize data following the real domain distribution, thus failing to generalize to challenging domains. Our proposed method addresses this limitation. 

\subsection{Ablation Studies}
In this section, we discuss the impact of various factors on the performance of our proposed DiffDFKD.
\paragraph{\textbf{The Effect of Different Diffusion Models}}
To assess the impact of various diffusion models on our proposed method, we conduct experiments primarily focusing on three diffusion models: DiT~\cite{peebles2022scalable}, GLIDE~\cite{nichol2021glide}, and Stable Diffusion (SD)~\cite{rombach2022high}. The experimental results, presented in Table~\ref{tbl:ablation_diffusion}, indicate that DiT and GLIDE outperform SD. This disparity in performance can be attributed to DiT and GLIDE being trained on the ImageNet dataset, whereas SD has not undergone such training. Despite the performance differences, it is important to note that our method utilizes SD. These results demonstrate that while the choice of diffusion model can influence performance, our proposed method is not inherently reliant on any single diffusion model and maintains effectiveness, showcasing its robustness and adaptability across various DMs.

\begin{table}[]
\centering
\caption{Ablation studies on diffusion model-based methods. We use ResNet34 as the teacher to train ResNet18 student model.}
\vspace{-2mm}
\begin{tabular}{@{\hspace{2pt}}lcl@{\hspace{2pt}}}
\toprule
Method & Syn Images Numbers & Acc (\%) \\
\midrule
Ours w/GLIDE & 3000 & 87.62 \\
Ours w/SD    & 3000 & 86.65 \\
Ours w/DiT   & 3000 & 88.98 \\
\bottomrule
\end{tabular}
\vspace{-3mm}

\label{tbl:ablation_diffusion}
\end{table}

\paragraph{\textbf{Latent Cutmix Augmentation}} 
In this section, we explore various data augmentation techniques~\cite{shorten2019survey} applied to the latent representations during the denoising process of the diffusion model. The ablation results in Table~\ref{tbl:ablation_cut_mix} show that while traditional~\cite{xu2023comprehensive, peng2024lightweight} and MixUp~\cite{zhang2017mixup} augmentations on the latents lead to a performance drop compared to using the raw latents, CutMix augmentation improves performance.

While MixUp and traditional augmentation methods introduce diversity, they cause the diffusion model's generative process to deviate from the convergence defined in~\eqref{eqn:unified}, resulting in increased loss. In contrast, our LCA approach operates on small regions between two latent representations, increasing diversity while ensuring the generated data adheres to DFKD criteria. 

\begin{table}[]
\small
\centering

\caption{ Student accuracy with different Latent Augmentation.}
\vspace{-1mm}
\begin{tabular}{ccrrrr}
\toprule
&  & \multicolumn{4}{c}{Latent Augmentation}                  \\
\multirow{-2}{*}{Teacher} & \multirow{-2}{*}{Student} & \multicolumn{1}{c}{None} & \multicolumn{1}{c}{Traditional} & \multicolumn{1}{c}{Mixup} & \multicolumn{1}{c}{Cutmix} \\
\midrule
R34  & R18  & \multicolumn{1}{c}{84.89} & \multicolumn{1}{c}{83.43} & \multicolumn{1}{c}{83.65} & \multicolumn{1}{c}{\textbf{86.56}} \\
V11   & R18 & \multicolumn{1}{c}{80.68} & \multicolumn{1}{c}{77.61} & \multicolumn{1}{c}{80.10} & \multicolumn{1}{c}{\textbf{81.59}} \\
W40-2& W16-1 & \multicolumn{1}{c}{81.11} & \multicolumn{1}{c}{79.50} & \multicolumn{1}{c}{80.63} & \multicolumn{1}{c}{\textbf{82.47}} \\
W40-2& W16-2 & \multicolumn{1}{c}{85.40} & \multicolumn{1}{c}{85.24} & \multicolumn{1}{c}{85.76} & \multicolumn{1}{c}{\textbf{87.01}} \\
W40-2& W40-1 & \multicolumn{1}{c}{85.01} & \multicolumn{1}{c}{83.62} & \multicolumn{1}{c}{84.49} & \multicolumn{1}{c}{\textbf{86.40}} \\
\bottomrule        
\end{tabular}
\vspace{-7mm}

\label{tbl:ablation_cut_mix}
\end{table}

\section{Conclusion}
 In this paper, we propose DiffDFKD, a novel framework that leverages pre-trained diffusion models for data synthesis and effectively utilizes teacher model knowledge to bridge domain gaps in DFKD. Our DiffDFKD guides the diffusion model's synthesis process using teacher model information, generating domain-customized data closely mirroring the original distribution. Additionally, we introduced Latent CutMix Augmentation, a computationally efficient technique that enhances the diversity of synthesized images while preserving essential characteristics for effective knowledge transfer. Experiments validated the efficacy of DiffDFKD, yielding state-of-the-art DFKD performance. The framework demonstrates the potential of harnessing diffusion models for DFKD.

\section*{Acknowledgment}
This work was supported by the Natural Science Foundation of China (62276242), National Aviation Science Foundation (2022Z071078001), Dreams Foundation of Jianghuai Advance Technology Center (2023-ZM01Z001).

\bibliographystyle{IEEEbib}
\bibliography{icme2025references}

\end{document}